%% file: main.tex
\def\year{2022}\relax
\documentclass[letterpaper]{article} 
\usepackage{aaai22}  
\usepackage{times}  
\usepackage{helvet}  
\usepackage{courier}  
\usepackage[hyphens]{url}  
\usepackage{graphicx} 
\usepackage{natbib}  
\usepackage{caption} 
\DeclareCaptionStyle{ruled}{labelfont=normalfont,labelsep=colon,strut=off} 
\usepackage{amsmath,amssymb,amsfonts}
\usepackage{algorithm}
\usepackage{algorithmicx}
\usepackage{algpseudocode}
\usepackage{makecell}
\usepackage{svg}
\usepackage{multirow}
\usepackage{newfloat}
\usepackage{listings}
\floatstyle{ruled}
\newfloat{listing}{tb}{lst}{}
\floatname{listing}{Listing}
\graphicspath{ {./images/} }
\title{Landmark Policy Optimization for Object Navigation Task}
\author{
    Aleksey Staroverov,\textsuperscript{\rm 1,2}
    Aleksandr I. Panov\textsuperscript{\rm 1,2}
}
\affiliations{
    \textsuperscript{\rm 1}Moscow Institute of Physics and Technology\\
   \textsuperscript{\rm 2}Artificial Intelligence Research Institute\\
   Moscow, Russia\\
   staroverov.av@phystech.edu
}

\begin{document}

\maketitle

\begin{abstract}
This work studies object goal navigation task, which involves navigating to the closest object related to the given semantic category in unseen environments. Recent works have shown significant achievements both in the end-to-end Reinforcement Learning approach and modular systems, but need a big step forward to be robust and optimal. We propose a hierarchical method that incorporates standard task formulation and additional area knowledge as landmarks, with a way to extract these landmarks. In a hierarchy, a low level consists of separately trained algorithms to the most intuitive skills, and a high level decides which skill is needed at this moment. With all proposed solutions, we achieve a 0.75 success rate in a realistic Habitat simulator. After a small stage of additional model training in a reconstructed virtual area at a simulator, we successfully confirmed our results in a real-world case. 
\end{abstract}

\section{Introduction}

Autonomous navigation in a semantically extensive environment is one of the significant components in building intelligent robotic systems. In this article, we set the navigation task in the form of an object goal navigation problem (ObjectNav). The agent (robot) appears in the previously unseen environment and has an RGBD camera and odometry sensors that measure the agent's position relative to the start of the episode. The episode is considered successful if the agent navigates to a place and executes the stop action at a small distance from the object related to the given semantic category.

\begin{figure}[ht!]
  \centering
  \includegraphics[width=10cm,height=10cm]{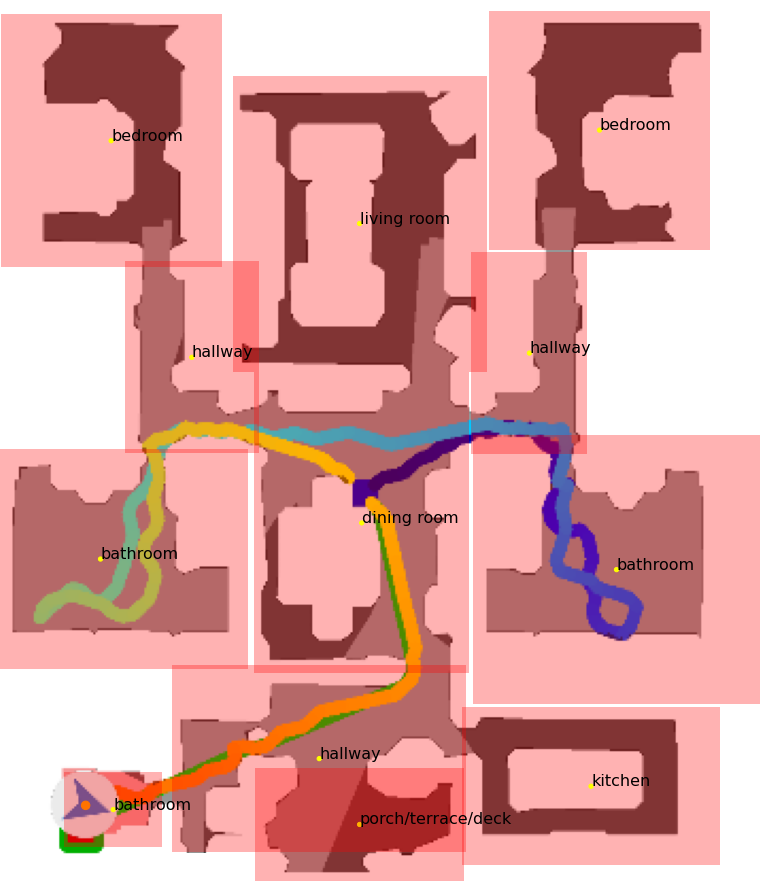}
  \caption{Figure showing a sample trajectory with the visualization of landmarks. }
  \label{fig:traj_map}
\end{figure}

Classically, such problems are solved with planning and simultaneous localization and mapping (SLAM) methods \cite{slam1}. As a result, the agent generates a collision-free path to the goal. If the goal object's coordinates are unknown, methods such as a Frontier-based exploration(FBE) \cite{fbe} are often used. A frontier is defined as the boundary between the explored free space and the unexplored space. Frontier-based exploration essentially samples points on this frontier as goals to explore the space.

Another way is to use the end-to-end Reinforcement Learning (RL) approach \cite{Sutton} \cite{ppo} \cite{sac} \cite{openai2019dota}. In general, reinforcement learning uses the idea of an agent that interacts with an environment. This interaction is formally described by the Markov decision process(MDP). During the training, the agent learns the policy that maps observation to a distribution over actions by the reward given from the environment. DDPPO \cite{ddppo} made a breakthrough and solved point navigation tasks at various simulated photorealistic scenes with an almost perfect score. Though it has some limitations, some of them, like the GPS sensor's presence, was solved by works that continue this approach. The ObjectNav task also remained too hard to that moment even with the additional semantic segmentation module, partially because it is not entirely clear how to determine the appropriate reward function.

Using a learnable map module and dividing a policy into a global, that by planning on a map output a short-term subgoal, and a local policy, that pursues that subgoal, SemExp \cite{semexp} has shown the best result at ObjectNav during Habitat Challenge 2020. At the 2021 challenge, end-to-end RL back to state-of-the-art (SOTA). The authors \cite{auxrl} achieved that by adding auxiliary learning tasks and an exploration reward. 

Significantly increasing metrics further is not possible due to unsolved episodes having too many areas to explore or semantically differs from others (datasets can contain scenes from private houses with open terraces to office space \cite{Matterport3D}). As humans solve exploration tasks to find the object, it strongly relies on a room's understanding of a concrete scene and can predict the type of objects inside. 

\begin{figure}[ht!]
    \centering
    \includegraphics[width=4cm]{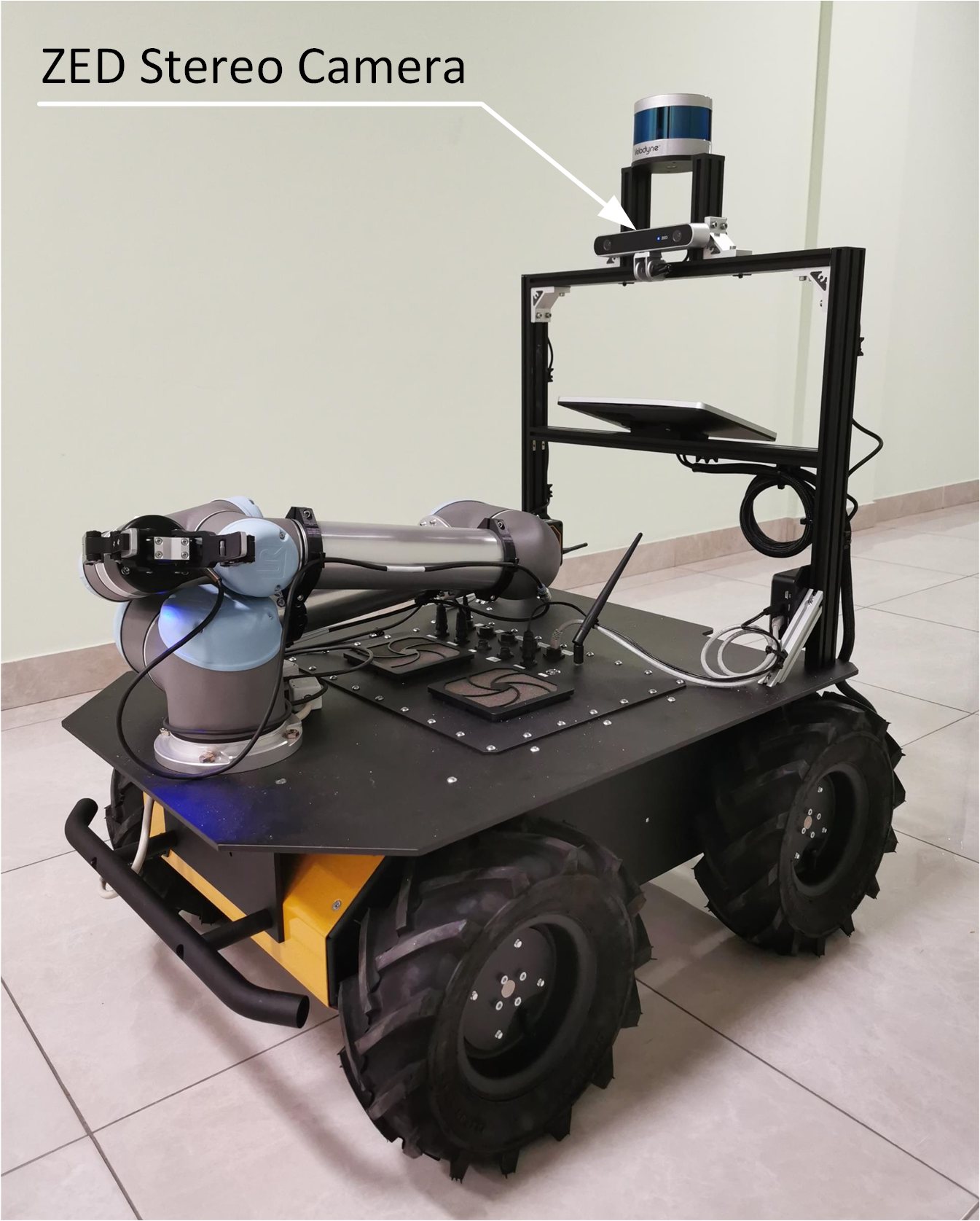}
    \caption{Ground robot platform based on the Clearpath Husky chassis with a ZED camera. We used it to evaluate our results in real-world scenarios.}
    \label{fig:husky}
\end{figure}

To solve these problems, our contribution is:
\begin{itemize}
\item \textbf{Task formulation with landmarks}.
To allow the agent to learn this concept, we gave to the agent the landmarks in a form of all rooms center coordinates and their kind.Note that obstacle maps or objects inside any of the rooms are still unknown to the agent. Our proposed method is to analyze this given list of landmarks and navigate to them one by one according to the global policy; if the goal-type object is not at the landmark area, navigate to the following most relevant; if in it, navigate directly to the object and complete the episode. 
\item \textbf{Dividing policy into a set of skills}. As an agent's policy, we distinguished three basic skills: navigation to the point, exploration of the nearby area, and reaching the seen object. The agent was separately trained in all of these skills using the RL policy. 
\item \textbf{Hierarchical structure}.
Then we combined these learned skills in a hierarchical way to the final agent behavior that solves the object goal navigation task. 
\item \textbf{Smooth policy transfer to new real-world scenes}. We 3D reconstructed our laboratory scene with a professional Leica RTC360 scanner to make the policy transfer possible and predictable. This allowed us to adopt our neural networks to a new location before actual tests at our Husky robot (Fig. \ref{fig:husky}). Also, as the Husky robot relies on an RGB camera, we used an additional depth reconstruction module to get an RGBD image.

\end{itemize} 

With this, we increase the episode success rate up to 75\% compared to the 50\% at the current SOTA method \cite{auxrl} due to the exploration of the scenes became more focused and semantically meaningful.
  
An example of episode trajectory can be seen in Fig. \ref{fig:traj_map}. The agent has a goal to navigate to a sink. The global policy module determines that it should be at a bathroom, and there are three bathrooms given as input information for this scene. Agent navigating to the closest bathroom to itself, explore it and proceed to the next until it finds the goal type of object at the last bathroom and successfully reaches it.

\begin{figure*}[h]
  \centering
  \includegraphics[scale=1.0,height=6cm,width=18cm]{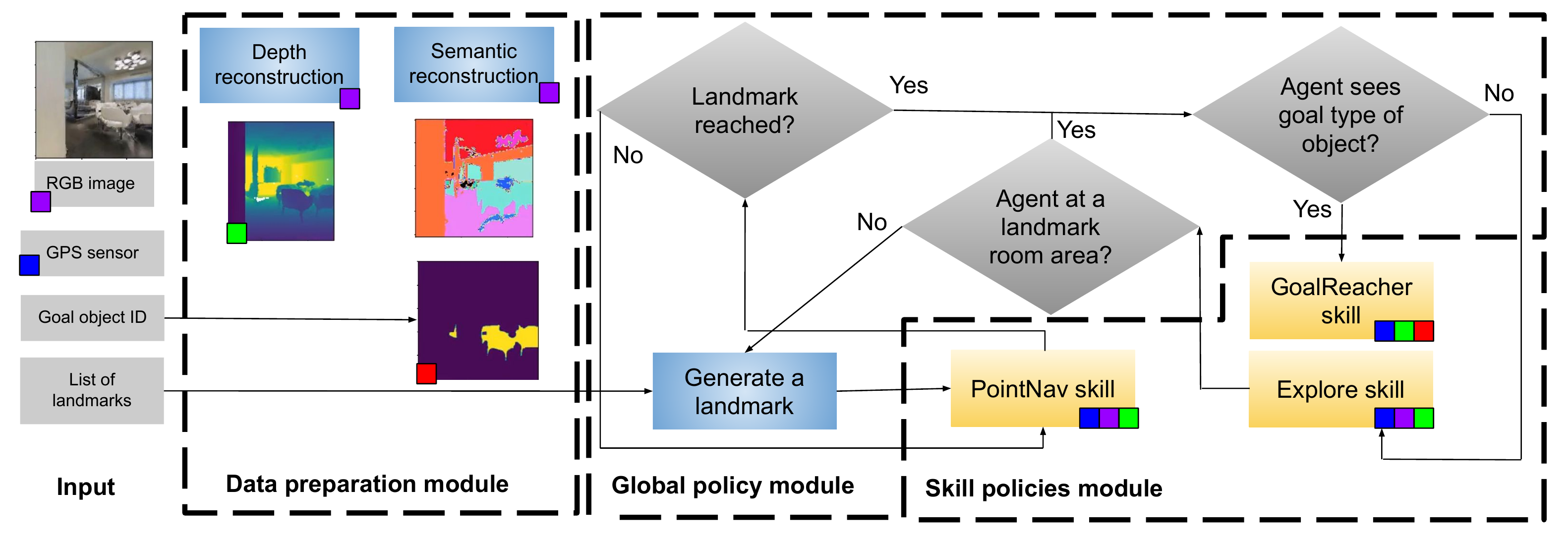}
  \caption{Hierarchical landmark policy optimization (HLPO) scheme. Our proposed approach consists of three main blocks: data preparation, global policy, and skill policies. Multicolored squares at the bottom of the elements mean what data it denotes (at the bottom left) and consume (at the bottom right).}
  \label{fig:hlpo_scheme}
\end{figure*}

\section{Related Work} 

Usually, humans treat navigation as one complex task, including SLAM,  planning a route on a map, setting subgoals, and interacting with the environment. In the classical approach, such a complex system is implemented through separate modules, each responsible for its task. Another possible pipeline is to use end-to-end RL algorithms. RL-based algorithms have presented massive progress in a couple of years, ranking first in different navigation challenges. Despite that, it could be computationally hard to train from the ground and struggle when deployed in previously unseen realistic large environments. 

A possible solution to deal with this problem is to simplify the environment to a 2D representation as a map or an occupancy grid \cite{2_sim} \cite{3_sim}. A trained model with SLAM or a laser scan input can then be transferred to a real robot. Despite that, when noises in sensors or external conditions of the real-world do not allow building a 2D representation with sufficient accuracy or the scene could not be treated as flat to navigate without collisions, there are many cases.

At the other side, more robust to noises end-to-end reinforcement learning with an appropriate training time demonstrates the best performance in some tasks. Without any mapping or planning modules, DDPPO at the PointNav task using 64 GPUs and three days of training with 2.5 billion steps in the environment reaches a success rate weighted by the path length (SPL metrics) equal to 0.997 and achieves human-like performance \cite{ddppo}. To optimize the training process, the authors investigated the most efficient neural network architecture \cite{ddppo_budget}.

To reduce training time and perform more complex tasks like ObjectNav, Hierarchical Reinforcement Learning (HRL), which enables autonomous decomposition of challenging long-horizon decision-making tasks into simpler subtasks, can be used. HRL could be formalized as global and local policies. In sparse rewards settings, a methods with two levels of policies was demonstrated by this works \cite{anm} \cite{semexp} \cite{ieee_access}. 

Another way to think about the global and local policy is to treat them as topological and metric spatial reasoning. Semantic instructions or contextual clues from the global policy need to be converted by the local policy into metrically precise terms to control the robot. In one such study \cite{4_nomap}, the authors use key points, as designed for humans, in human-made environments, such as airports, to build planning algorithms, rather than relying on geometric maps.

Real realistic data in the fast-performing simulator is needed for the RL algorithm's training process and its further transfers to the real world. In the past few years, recent advancements in annotated 3D maps of the real-world data have appeared in the form of 3D reconstructed spaces datasets, such as Stanford2D3DS \cite{stanford2d3d} and Matterport3D \cite{Matterport3D}. As for the simulator to load this datasets, we choose from  MINOS \cite{minos},  Gibson \cite{gibson}, Habitat \cite{habitat}, and THOR \cite{ai2thor}. We decided to use the Matterport3D dataset because of its size and diversity and the Habitat environment because of its rendering speed and straightforward way to multithread. To train PointNav policy that does not require a semantic sensor, we used the faster version of Habitat (that does not support native semantic segmentation sensor), the BPS simulator \cite{bps}, which was 100x timed faster. With a similar approach, these works  \cite{sim2realhabitat} \cite{iros_panoramic_task}  showed the ability to transfer the RL model from a simulated environment to real-world usage.

\section{Task Setup} 

In its simplest form, the indoor object navigation task is defined as the task of navigating to an object (specified by semantic label) in a previously unseen environment \cite{objectnav_task_formulated}. In practice, the agent is initialized at a random pose in an environment and aims to find an instance of an object category $C={c_{1},c_{2},...,c_{20}}$ (for example, a \textit{couch}) by navigating to it. This interaction is formally described by the Markov decision process (MDP), which is defined by sets of states ${S}$ and actions ${A}$ (\textit{forward}, \textit{turn left}, \textit{turn right}, and \textit{stop}), the distribution of the initial states ${p}({{s}_{0}})$, the reward function ${r}:{S}\times{A}\to\mathbb{R}$, the transition probabilities ${p}({s}_{t+1}\mid{s}_{t},{a}_{t})$, the termination probabilities ${T}({s}_{t},{a}_{t})$, and the discount factor $\gamma\in[0,1]$. The agent receives a semantic mask of a goal-type of the object through the semantic segmentation module $\Phi_{semantic}$. The scene map is not available to the agent. During the evaluation process, the agent can only use the input from the RGB-D camera, the GPS+Compass sensor, and a list of landmarks ($\mathcal{G}$) for navigation. The GPS+Compass sensor provides the agent’s current location and the orientation information relative to the start of the episode. A list of landmarks contains all center coordinates of rooms and their type that are in the scene with no information about the map, what objects are inside, or how to navigate to those rooms.

Evaluation occurs when the agent selects the \textit{stop} action. As a metric, the Success rate weighted by Path Length (SPL) and the Success rate are used. SPL is computed to the object instance closest to the agent start location. 
\begin{equation}\label{spl}
SPL = \frac{1}{N}\sum_{i=1}^{N}  \frac{l_{i}}{max\left(p_{i},l_{i}\right)}
\end{equation}
where $l_i$ is the length of shortest path between the goal and the target for an episode, $p_i$ is the length of the path taken by the agent in an episode.

Thus, if an agent spawns very close to \textit{chair1} but stops at a distant \textit{chair2}, it will achieve 100\% success (because it found a ``chair") but a fairly low SPL (because the agent path is much longer compared to the true path). More specifically, an episode is deemed successful if the agent is calling the \textit{stop} action within 1.0m Euclidean distance from any instance of the target object category, and an oracle can view the object from that stopping position by turning the agent or looking up/down.

\section{Methods} 

We propose a landmark-based modular framework (Fig. \ref{fig:hlpo_scheme}) for navigation to object goal, ``Hierarchical landmark policy optimization" (HLPO). The framework consists of three main modules: global policy $\pi_{global}$, data preparation $\Phi_{semantic}$, and skill policies $\{\pi_{explore},\pi_{reacher},\pi_{pointnav}\}$. 

\begin{algorithm}
\caption{HLPO}\label{alg:policy_test}
\begin{algorithmic}[1]
\scriptsize
\State \textbf{Given:\hfill \break} 
        $\pi_{global}:$ Landmark policy,\hfill \break
        $\pi_{explore}:$ Explore policy,\hfill \break
        $\pi_{reacher}:$ GoalReacher policy,\hfill \break
        $\pi_{pointnav}:$ PointNav policy,\hfill \break
        $\Phi_{semantic}:$ Semantic segmentation model,\hfill \break
        
\State \textbf{Input:\hfill \break} 
        $RGBD:$ RGBD image from the camera,\hfill \break
        $GPS:$ X,Y coordinates relative to the start point,\hfill \break
        $\mathcal{G}:$ List of landmarks.\hfill \break  
        
\While{episode episode not ended}
    \State $room\_cord \gets \pi_{global}(\mathcal{G},goal\_type,GPS)$
    \While{The agent is not inside a landmark area.}
        \State $a \gets \pi_{pointnav}(RGBD,GPS,room\_cord,a_{prev})$
        \State Execute action $a$ in the environment
    \EndWhile
    \State $To\_goal=False$
    \State $semantic \gets \Phi_{semantic}(RGB)$
    \While{The agent is inside a landmark area.}
        \If{$ \overline{To\_goal}:$} \label{lst:line:reacher}
            \State $a \gets \pi_{reacher}(Depth,semantic,GPS,a_{prev})$
        \Else
            \State $a \gets \pi_{explore}(RGBD,GPS,a_{prev})$
        \EndIf
        \State Execute action $a$ in the environment    
        \State $semantic \gets \Phi_{semantic}(RGB)$
        \If{$semantic \not={0}$}
            \State $To\_goal \gets True$
        \EndIf
    \EndWhile    
\EndWhile    
\end{algorithmic}
\end{algorithm}

\begin{figure*}[h]
\minipage{0.32\textwidth}
  \includegraphics[scale=0.4,height=5cm,width=6cm]{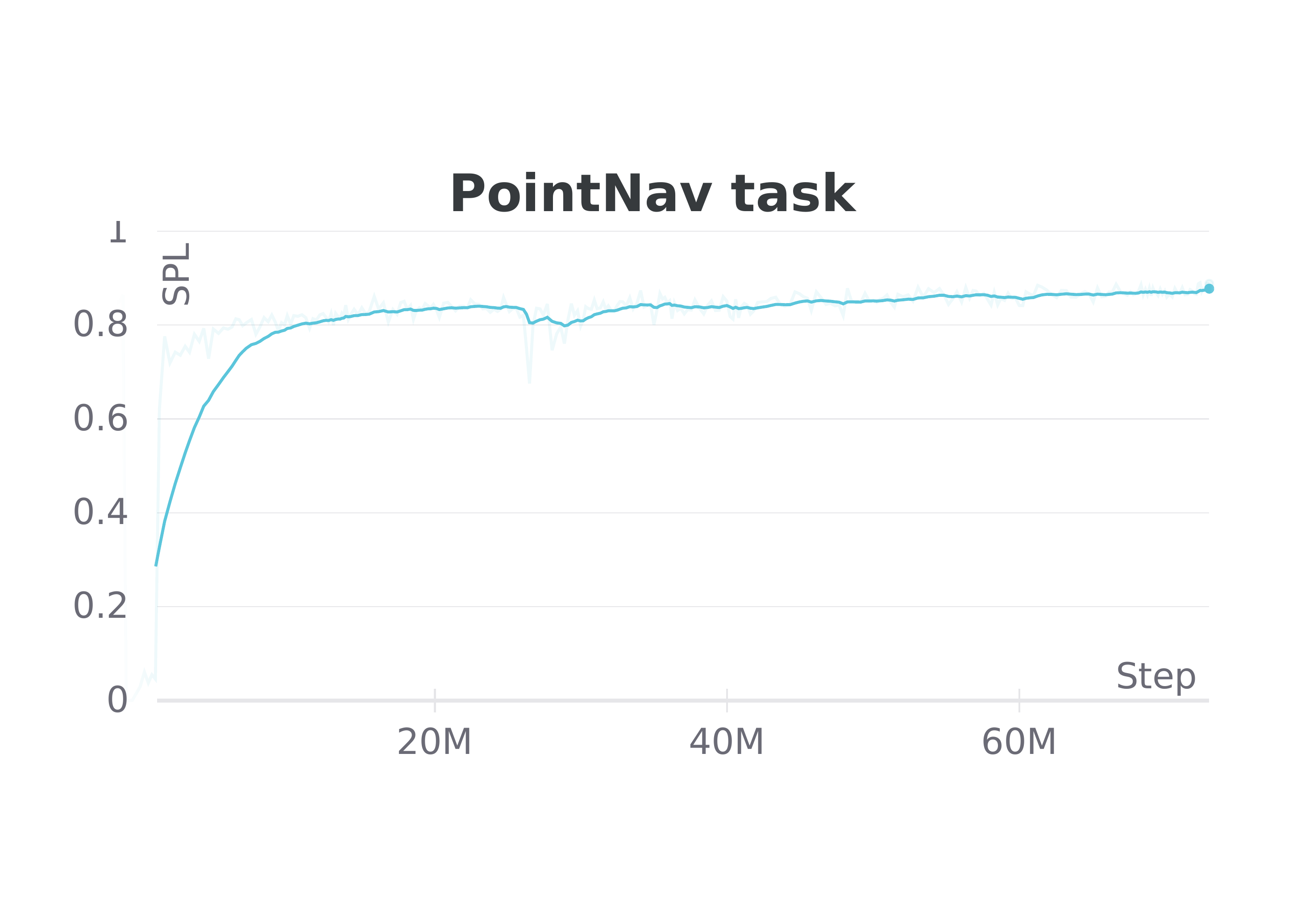}
  \caption{Success rate weighted by Path Length (SPL) vs. steps on a training phase of PointNav skill.}
  \label{fig:pnav_wb}
\endminipage\hfill
\minipage{0.32\textwidth}
  \includegraphics[scale=0.4,height=5cm,width=6cm]{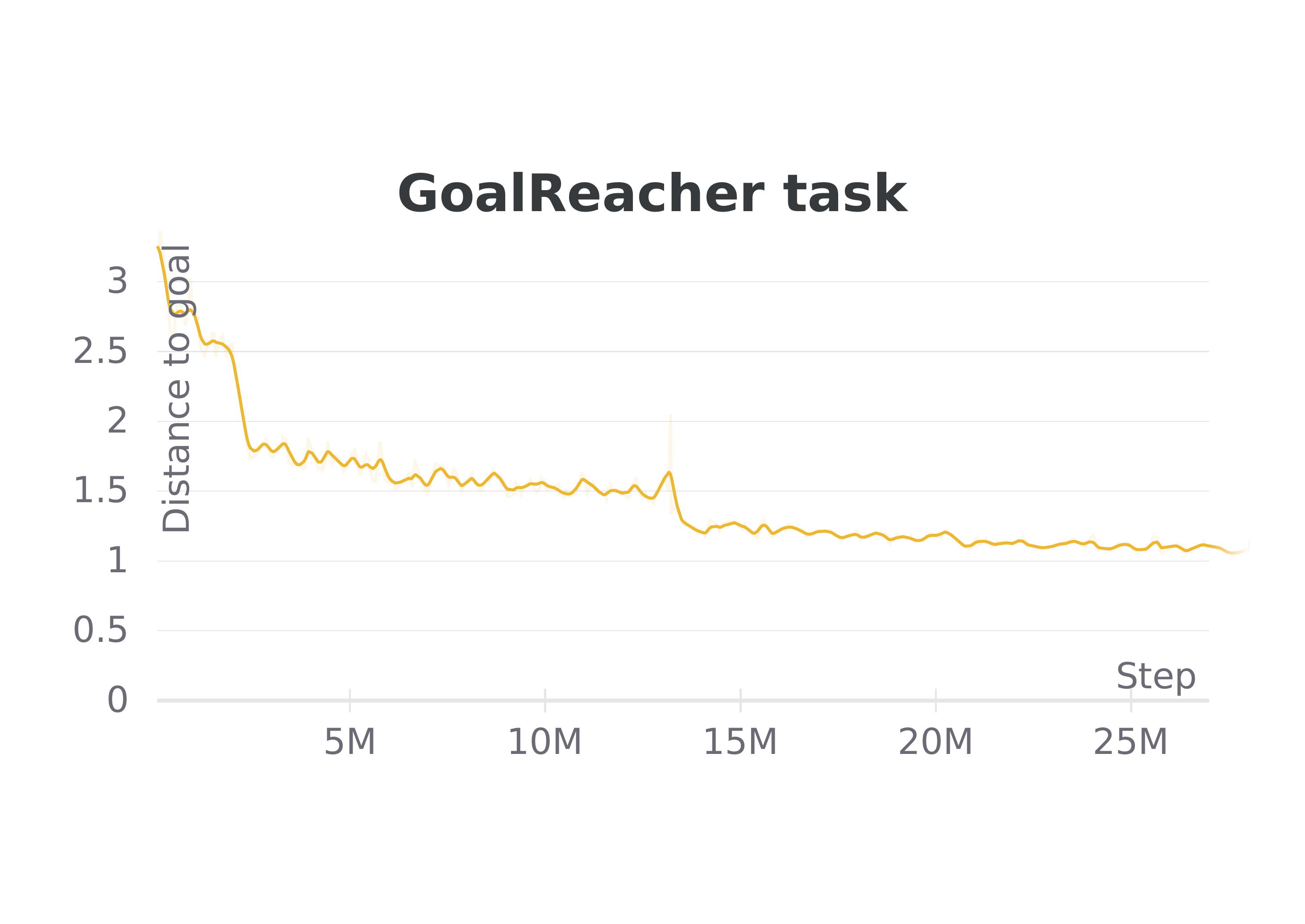}
  \caption{Distance to the goal type of object ($m$) vs. steps on a training phase of GoalReacher skill.}
  \label{fig:onav_wb}
\endminipage\hfill
\minipage{0.32\textwidth}%
  \includegraphics[scale=0.4,height=5cm,width=6cm]{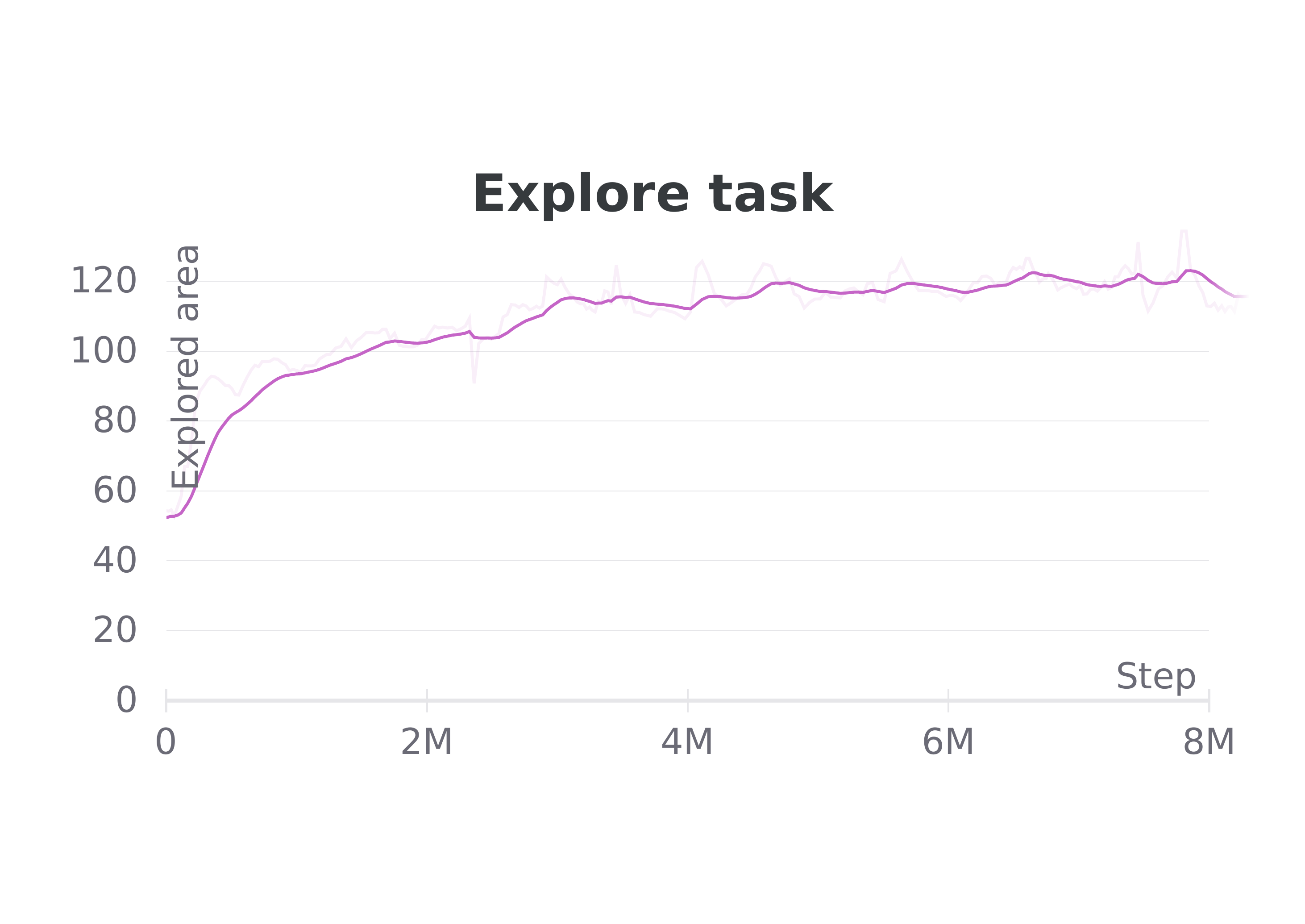}
  \caption{Explored area ($m^{2}$) vs steps on a training phase of Exploration skill.}
  \label{fig:enav_wb}
\endminipage
\end{figure*}

\textbf{The global policy module} analyze all training scenes to connect the object types to the type of rooms and made a statistic (Fig.3 in supplementary materials). Based on this statistic, the current object goal type and distances to rooms at the current scene, global policy module ranging rooms (as the probability of finding the object in a room/distance to room) in order of needs to visit to find a goal object.

\textbf{The data preparation module} is two neural nets that do semantic segmentation and depth reconstruction. Though we use a depth sensor from a simulator for all training and testing at the simulator, we do not have that option at the Husky robot, so we reconstructed it from an RGB image.

\textbf{The skill policies module} takes as input agents observation and outputs the action that pursues current needed skill. For our task, there are three skills: point navigation (PointNav, Fig. \ref{fig:pnav_wb}), exploration (fig. \ref{fig:enav_wb}), and goal reacher (GoalReacher, Fig. \ref{fig:onav_wb}). 

The PointNav skill was trained as a subtask to reach a landmark (room of interest) in those scenarios. We have test two versions of it, RL and Fast Marching Method (FMM). Even at complex labyrinth-like areas with high geodesic distance, the RL version turns out to have better results than a SLAM with a planning method.

The GoalReacher skill was trained to reach the object goal when the semantic sensor sees it (Algorithm \ref{alg:policy_test} line~\ref{lst:line:reacher}). Its exploration of the scene abilities are low, but at short distances, it avoids obstacles to the goal well, and it has a good sense of when the seen object is reached within a given precision and the episode needs to stop. RL is essential at this task because semantic segmentation could provide a lot of noise. If we draw at least one wrong prediction to the map, we could end up at this wrong spotted place, while RL does not directly reconstruct the semantic map and is trained with the presence of these noises and can avoid them well.

The Exploration skill determines the task's success more than others, especially when no landmarks are given. We trained an RL policy that effectively explores the nearby area, so it fits perfectly to explore the room completely but has lower percentage coverage at big scenes.

The process of managing skills could be treated as a policy and integrated into the global policy module. Typically, the HRL methods can learn a k-level policy ${\Pi}_{k-1}$. Each level of policy learns ${\pi}_{i}:{S}_{i},{G}_{i}\to {A}_{i}$ where $G_i$ is the set of possible sub-goals. To learn these policies ${\pi}_{i}$ the set of MDPs ${U}_{0}$,${U}_{1}$, in which ${U}_{k}=(S,G,A,T,R,\gamma)$ are used. However, learning multiple levels of policies in parallel is problematic because it is inherently unstable. For the object goal task, the sequence of skills can be formulated explicitly. The agent navigates to the first landmark (coordinate the room of interest given by global policy) by point navigation skill. Then the agent explores the room until it leaves it by exploration skill. If the semantic segmentation model sees a goal type of the object in the room of interest, the goal teacher skill activated and navigated to this object (Algorithm \ref{alg:policy_test}).

All skill policies in our method are using RL. In general, RL consists of an agent that interacts with an environment. A policy ${\pi}$ of the agent is a function that maps a state to a distribution over actions. The episode starts with the initial state ${s}_{0}$. At every step ${t}$, the agent executes the action given by a policy ${a}_{t}\sim\pi(\cdot\mid{s}_{t})$. As a response, the environment returns a reward ${r}={r}({s}_{t},{a}_{t})$ to the agent. With probability ${T}({s}_{t},{a}_{t})$ the episode is terminated, if not, a new state ${s}_{t+1}$ of the environment is sampled from ${p}(\cdot\mid{s}_{t},{a}_{t})$. The discounted sum of future rewards (return) is defined as ${R}_{t}=\sum^{\infty}_{i=t}\gamma^{i-t}{r}_{i}$. The agent's goal is to find the policy $\pi$ that maximizes the expected return $\mathop{\mathbb{E}}_{\pi}[{R}_{0}\mid{s}_{0}]$). The expectation is taken over the initial state distribution, the policy, and the environment transitions accordingly to the dynamics specified above. The action-value function (${Q}$-function) of a given policy $\pi$ is defined as ${Q}^\pi({s}_{t},{a}_{t})=\mathop{\mathbb{E}}_{\pi}[{R}_{t}\mid{s}_{t},{a}_{t}]$. The state-value function (${V}$-function) is defined as ${V}^\pi({s}_{t})=\mathop{\mathbb{E}}_{\pi}[{R}_{t}\mid{s}_{t}]$. The advantage is defined as ${A}^\pi({s}_{t},{a}_{t})={Q}^\pi({s}_{t},{a}_{t}-{V}^\pi({s}_{t})$ and informs if action ${a}_{t}$ is better than the average action the policy $\pi$ takes in the state ${s}_{t}$.

De facto, the policy and the value functions are represented as two neural networks. The first represents the current policy $\pi$, and a value network approximates the current policy's value function ${V}\approx{V}^\pi$.

A policy neural network is a two-head network, one for the action distribution (actor) and the other for the action value estimation (critic). The actor stream is one FC layer that outputs logits for each out of four actions. An action to execute is picked as a categorical distribution of that logits. The critic stream is an FC layer that outputs value for the given state.

To compute the return, we use a generalized advantage  estimator (GAE) with $\gamma=0.99$ and $\tau=0.95$. Other hyper-parameters are listed (Table 1) in supplementary materials. The GAE is a method that combines multi-step returns in the following way:

\begin{equation}\label{gae_eq}
\delta^{V}(t)= r_{t}+\gamma V(s_{t+1}) - V(s_{t})
\end{equation}
\begin{equation}\label{gae_eq2}
\hat{A}^{GAE(\gamma,\lambda)}_{t}= \sum^{\infty}_{l=0}(\gamma\lambda)^{l}\delta^{V}_{t+l}
\end{equation}

For the policy loss function, we use the proximal policy optimization (PPO) \cite{ppo}, which still remains the SOTA solution in reinforcement learning at the vast quantity of tasks. The rewards are given proportionally to the reduction in distance to the closest goal instance. In our method, for training an RL agent, we use a gradient descent over a policy agent. Given a $\theta$-parameterized policy ${\pi}_{\theta}$ and a set of trajectories collected with it (commonly referred to as a ``rollout"), the agent updates ${\pi}_{\theta}$ as follows. Let $\hat{A}_{t}={R}_{t}-\hat{V}_{t}$, be the estimate of the advantage, where ${R}_{t}=\sum_{i=t}^{T}\gamma^{i-t}{r}_{i}$ and $\hat{V}_{t}$ is the expected value of ${R}_{t}$, and ${r}_{t}(\theta)$ be the ratio of the probability of the action under the current policy and the policy used to collect the rollout. The parameters are then updated by maximizing

\begin{equation}\label{ppo_eq}
J^{PPO}(\theta)= E_{t}\left[min(r_{t}(\theta)\hat{A}_{t} ,clip(r_{t}(\theta),1-\varepsilon,1+\varepsilon  )\hat{A}_{t})\right]
\end{equation}

As the parallelization method, we utilize the decentralized distributed proximal policy optimization (DD-PPO) way \cite{ddppo}. As a general abstraction, this method implements the following: at step $k$, worker $n$ has a copy of the parameters, $\theta_{n}^{k}$, calculates the gradient, $\partial\theta_{n}^{k}$, and updates $\theta$ via

\begin{equation}\label{ddppo_eq}
\theta_{n}^{k+1}= PU\left(\theta_{n}^{k},AR(\triangledown_{\theta}J^{PPO}(\theta_{1}^k),...,\triangledown_{\theta}J^{PPO}(\theta_{N}^k)) \right )
\end{equation}

where $PU$ is any first-order optimization technique (gradient descent), and $AR$ performs a reduction (mean) over all copies of a variable and returns the result to all workers.

\section{Experimental Setup} 

There are several ways to train agents to navigate, from training in real-world scenarios to fully simulated environments. The former is too inefficient as it needs a lot of resources and could bring a lot of harm while policy did not get optimal. The latter is comfortable working with, and outstanding results could be squeezed out, but our overall task is to navigate in real-world scenes. In this case, the transfer process from a simulated environment to an actual robot could be even more challenging than the training itself. In the simulator, the agent does not suffer from many real-world sensors imperfections. 

As an intermediate approach, we use a photorealistic environment where lidars and cameras 3D reconstruct the real-world scenes. All sensors are simulated and could be noise or limited, thus not being different from their prototypes. 

To make a transfer of the agent model smoother, we reconstructed our scene, proved that the agent succeeds in it, and only then brought our Husky robot to prove our method final.

\begin{figure*}[h]
  \centering
  \includegraphics[width=0.95\textwidth,height=9cm]{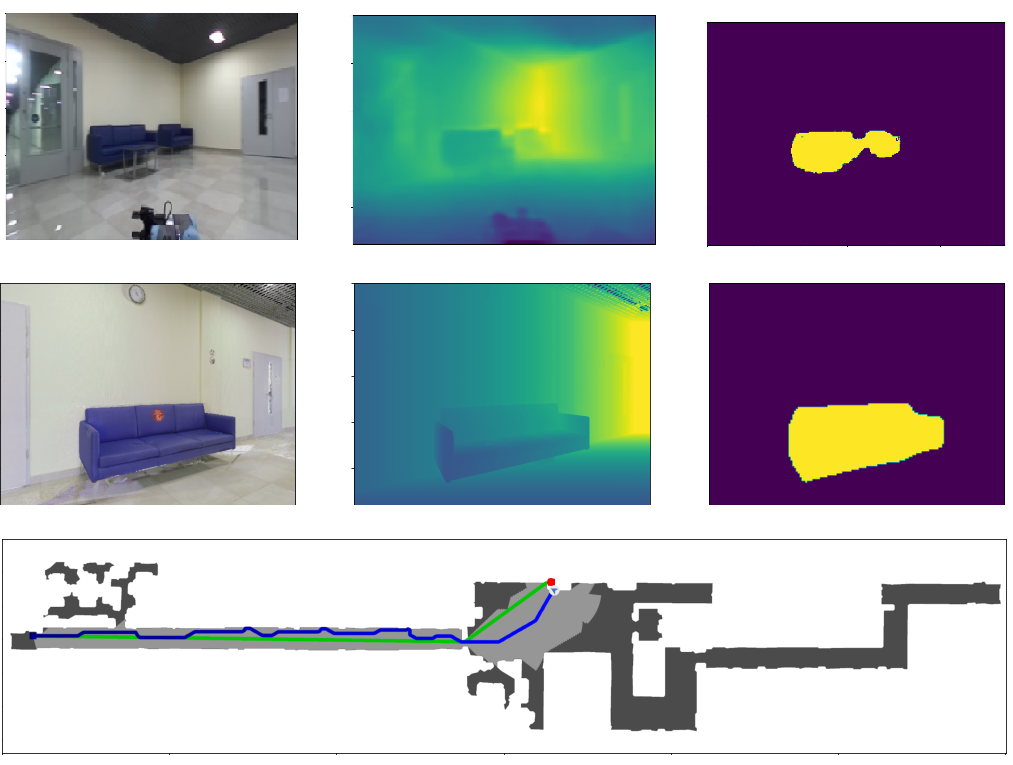}
  \caption{The bottom image is an example of what a reconstructed scene map looks like. The middle row shows how RGB,  depth, and semantic sensors at the Habitat simulator look. The top row is what the Husky robot gets at real tests. The data preparation module receives an RGB image and outputs the depth image and a semantic mask.}
   \label{fig:mipt_habitat}
\end{figure*}

\begin{table}[htbp]
  \centering
    \begin{tabular}{l|cccc}
        \hline
        \multirow{2}{*}{\bfseries Method} &
        \multicolumn{2}{c}{\bfseries GT semantic}&
        \multicolumn{2}{c}{\bfseries Learned semantic}\\
        
        & Success & SoftSPL & Success & SoftSPL\\ \hline
        E2E RL                  & 0.18 & 0.35 & 0.11 & 0.24\\
        SemExp                  & 0.24 & 0.26 & 0.11 & 0.17\\
        Planning                & 0.31 & 0.26 & 0.15 & 0.18\\
        Auxiliary RL            & 0.51 & 0.34 & 0.19 & 0.19\\
        2RL                     & 0.46 & 0.33 & 0.20 & 0.21\\

        \\ \Xhline{4\arrayrulewidth} \\
        HLPO (Plan)             & 0.68 & 0.43 & 0.37 & 0.30\\
        \textbf{HLPO}           & \textbf{0.75} & \textbf{0.38} & \textbf{0.46} & \textbf{0.30}\\
        HLPO (Map)              & 0.90 & 0.54 & 0.61 & 0.42 \\
        \hline 
    \end{tabular}
  \caption{ObjectNav results}
  \label{tab:results}
\end{table}   

To reconstruct the scene, we first looked at how it was done at the mp3d dataset. Dataset creators used the Matterport Pro2 camera (134 megapixels with no lidar) and the Matterport proprietary soft, where you can upload photos, and they are automatically processed to the final .obj file. That could be formatted to .glb and be used by the Habitat simulator with no effort. This works fine unless the scene contains small details. The mp3d dataset itself had a lot of holes and texture inconsistencies. We wanted higher quality, especially more precise depth reconstruction, so the agent can predictively navigate in a narrow room space. Our solution was to use a professional laser scanner Leica RTC360 3D. An additional plus is that we can manually edit our shots, delete background people, and manually check the quality of assembling the entire scene. To texture the final point cloud, we use a RealityCapture program (Fig.4 in supplementary materials). 

The example of an agent's trajectory is seen in Fig. \ref{fig:mipt_habitat}. We tested the agent from five different starting points within a 15m radius from the closest couch, and it does it successfully with SPL 0.8. The map was not available for the agent. It only uses an RGBD sensor, GPS relative to the start point, and a landmark as a coordinate at the lounge's center. The semantic mask was obtained by the semantic reconstruction module. As a depth net, we use \cite{depth_net}. As a semantic segmentation, we use SOLOv2 \cite{solov2} architecture.

To compare performance at a Habitat simulator, we executed the testing phase at test mp3d scenes at 100 episodes not presented during training. These episodes have a medium geodesic distance to the closest goal (Fig.1 in supplementary materials), not less than
5m and have a variety of goal objects to test the semantic model (Fig.2 in supplementary materials). We ran all experiments three times and got the dispersion of the results no more than 0.03.

\textbf{Baselines}. We use a range of methods as baselines: (Table~\ref{tab:results})

\begin{itemize}
  \item \textbf{E2E RL} - End-to-end DDPPO \cite{ddppo} algorithm trained at the depth + GPS + GT semantic sensors with the reward proportional to the geodesic distance to the goal object closest to the start. As a backbone, ResNet50 and LSTM layer were used.
  \item \textbf{SemExp} - A modular model that tackles the Object Goal navigation task \cite{semexp}. The SemExp is the SOTA algorithm with the module structure that incorporates both RL and planning, which showed the best performance at Habitat challenge 2020.
  \item \textbf{Planning} - A combination of planning modules that, at every step, update the obstacle map and explore it. When the semantic module sees the goal object, it gets spotted on a map and the agent navigates to it through planning.
  \item \textbf{Auxiliary RL} - The generic learned policy \cite{auxrl} with the auxiliary learning tasks and an exploration reward. It is the SOTA end-to-end algorithm, which showed the best performance at Habitat challenge 2021.
  \item \textbf{2RL} - A combination of two RL policies, one explores the area (Fig. ~\ref{fig:enav_wb}) until the semantic sensor sees the target and the other RL follows the seen goal object (Fig. ~\ref{fig:onav_wb}).
\end{itemize}

Other algorithms, as our main suggestion, need a piece of additional information about the environment. We provide to the agent all rooms center coordinates with no kind of map or information about what objects are inside

\begin{itemize}
  \item \textbf{HLPO (Plan)} - Close to HLPO, but instead of RL policies, we used planner on map modules. The map is also unknown from the start and gets reconstructed by the agent at each step.
  \item \textbf{HLPO} - Our proposed method uses PointNav RL (Fig. ~\ref{fig:pnav_wb}) to navigate the landmark area, Exploration RL (Fig. ~\ref{fig:enav_wb}) to explore the landmark area, and GoalReacher RL (Fig. ~\ref{fig:onav_wb}) to navigate to the goal object if the semantic module sees it.
  \item \textbf{HLPO (Map)} - To show maximum possible performance, we also gave the agent the obstacle map to get to the landmark with the FMM planning module as fast as possible. When inside a landmark area, two RL policies are acting the same as the native HLPO method. It should be noted that despite the given map, the agent still needs to explore it to find the semantic goal object. We assume that the condition of the predefined obstacle map to the agent is feasible to the real-world scenarios, but the semantic map with coordinates of all objects is not.
\end{itemize}

\section{Discussion and Conclusion}

We propose a novel approach to the ObjectGoal navigational task. With the standard formulation of the task, existing methods are limited to the point where, at large scenes, exploration without any information about the scene takes unreasonably much time. To solve this, we propose landmarks as a list of rooms coordinates and their type. These landmarks could gather for the scene without any additional lidars and computations. Also, because we do not incorporate an obstacle or semantic map, a scene could rebuild a lot, and items could be mixed in any order. As long as the locations of the rooms remain the same, there is no need to update landmarks information. 

With our updated task formulation, we have built a novel hierarchical policy that uses pretrained skills that could be stacked and reused in various navigational tasks without any changes. The success rate for our method doubles from 20\% for the state-of-the-art method to 46\% with the learned semantic and from 51\% to 75\% for the ground truth semantic from the simulator.  To accomplish the ObjectGoal task, we trained the agent three skills: PointNav, Exploration, and GoalReacher. We proved that RL is the best option for a policy for all agent skills as it is more robust to sensor noises and does not suffer from map reconstruction errors. Planning modules surpass RL and shows a close to perfect score with GT semantic only in almost ideal conditions or if the obstacle map of the scene is provided. 

To make the transfer to reality process possible and predictable, we described how to reconstruct a scene into a simulator with a decent photorealistic reconstruction quality using a professional Leica RTC360 scanner.

Future work plans to build a global policy that would automatically select pretrained skills for a more complex task and move from discrete actions to continuous ones to control wheel-based robots more effectively.


\bibliography{main}
\clearpage
\input{suppl}

\end{document}

%% file: suppl.tex
\section{Supplementary Material}

To compare performance at a Habitat simulator, we executed the testing phase at test mp3d scenes at 100 episodes not presented during training. These episodes have a medium geodesic distance to the closest goal (Fig.1), not less than
5m and have a variety of goal objects to test the semantic model (Fig.2).

\begin{figure}[htbp]
  \centering
  \includegraphics[width=8cm,height=5cm]{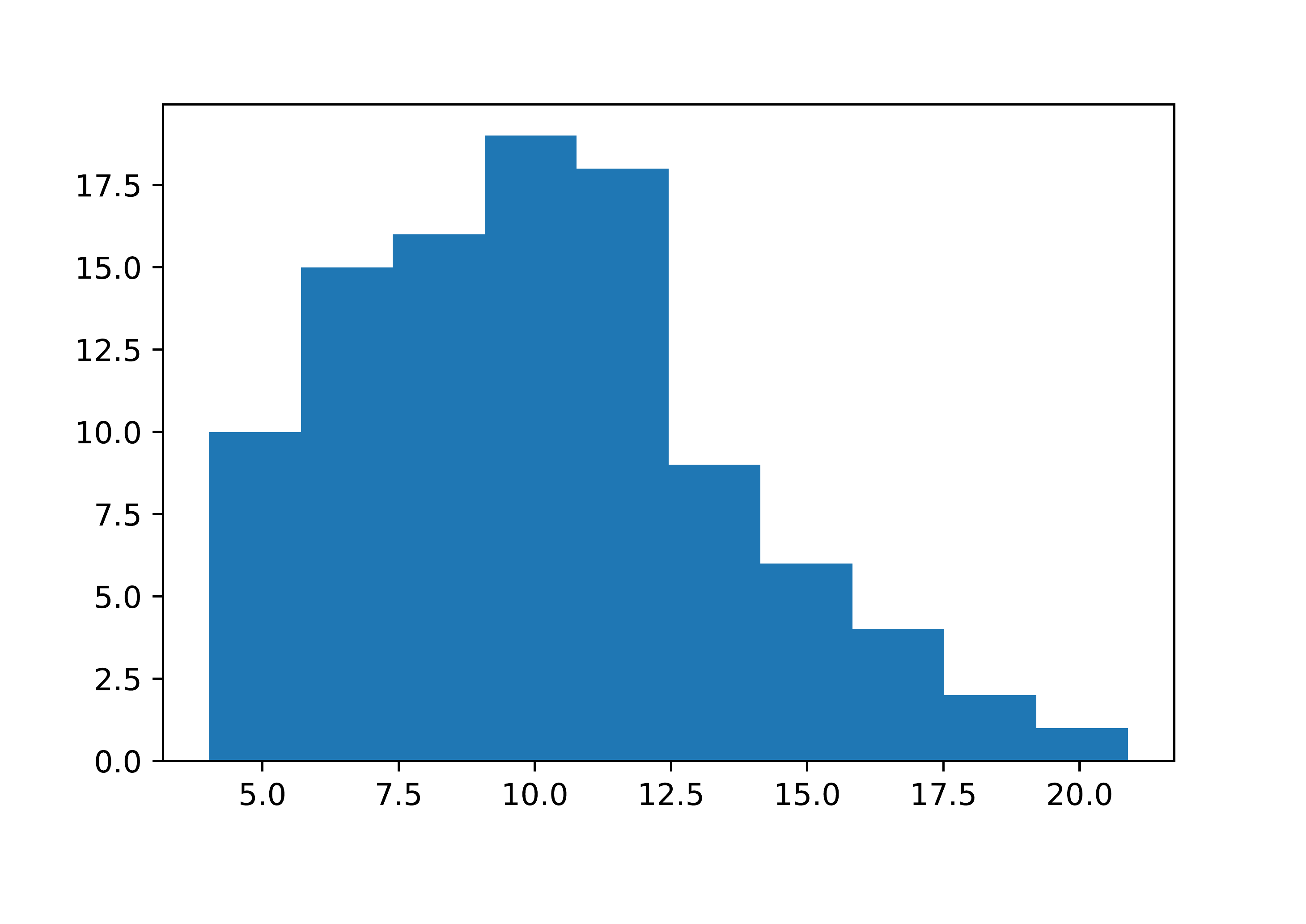}
  \caption{Histogram of distances to goal at test dataset.}
  \label{fig:epsdist}
\end{figure}  

\begin{figure}[htbp]
  \centering
  \includegraphics[width=6cm,height=5cm]{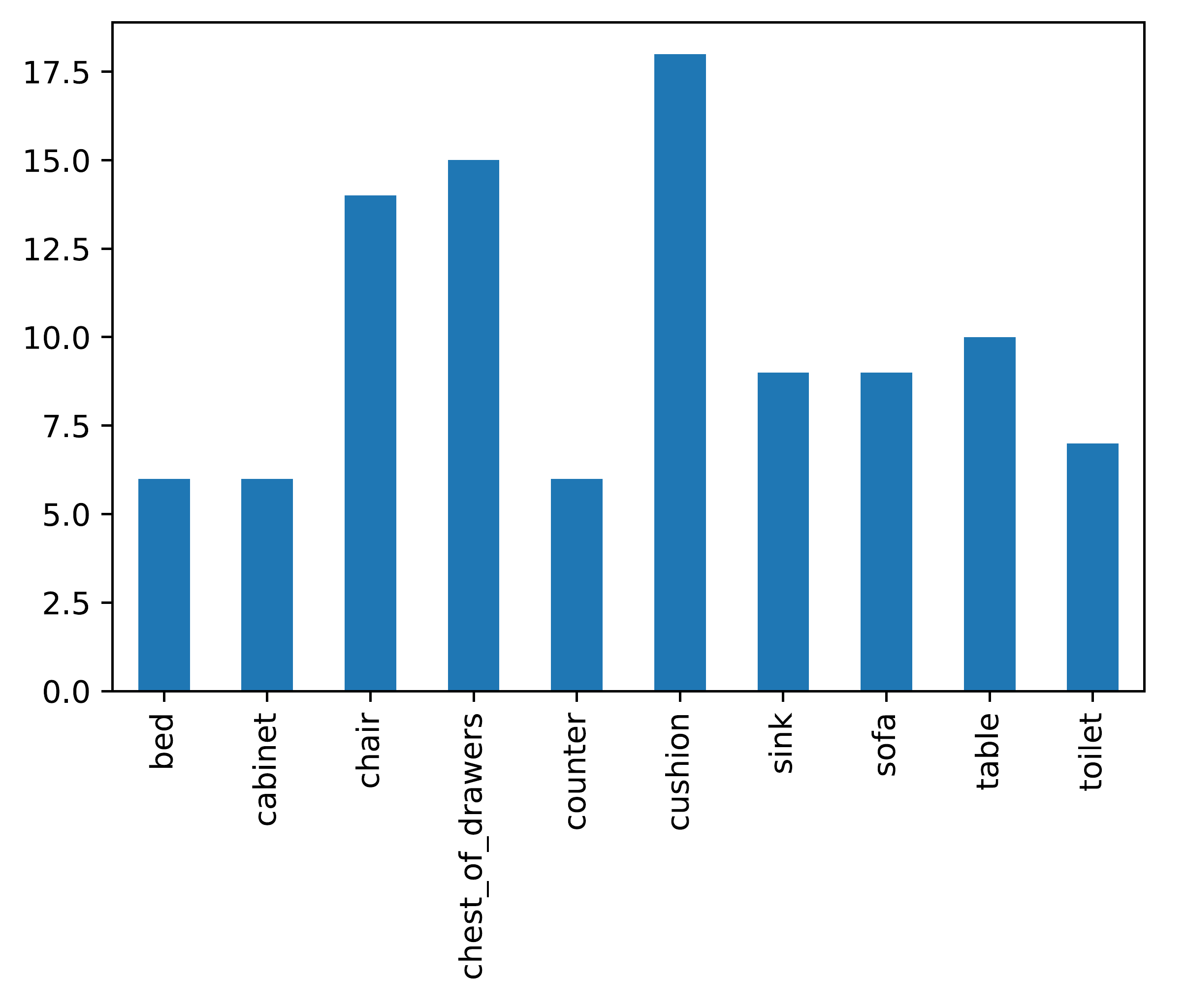}
  \caption{Types of the goal objects at test dataset.}
  \label{fig:objtypes}
\end{figure}

The global policy module analyze all training scenes to connect the object types to the type of rooms and made a statistic (Fig.3). Based on this statistic, the current object goal type and distances to rooms at the current scene, global policy module ranging rooms (as the probability of finding the object in a room/distance to room) in order of needs to visit to find a goal object.

\begin{figure}[htbp]
  \centering
  \includegraphics[width=8cm,height=22cm]{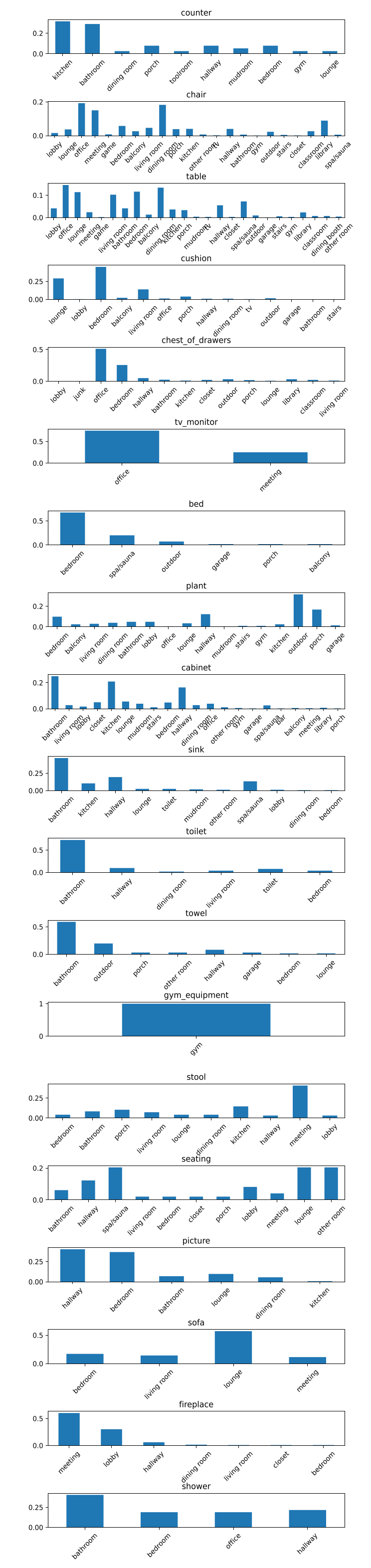}
  \caption{Room statistic}
   \label{fig:room_statistic}
\end{figure}

Our solution to reconstruct the scene was to use a professional laser scanner Leica RTC360 3D. An additional plus is that we can manually edit our shots, delete background people, and manually check the quality of assembling the entire scene. To texture the final point cloud, we use a RealityCapture program (Fig.4).

\begin{figure*}[h]
  \centering
  \includegraphics[width=0.95\textwidth,height=4cm]{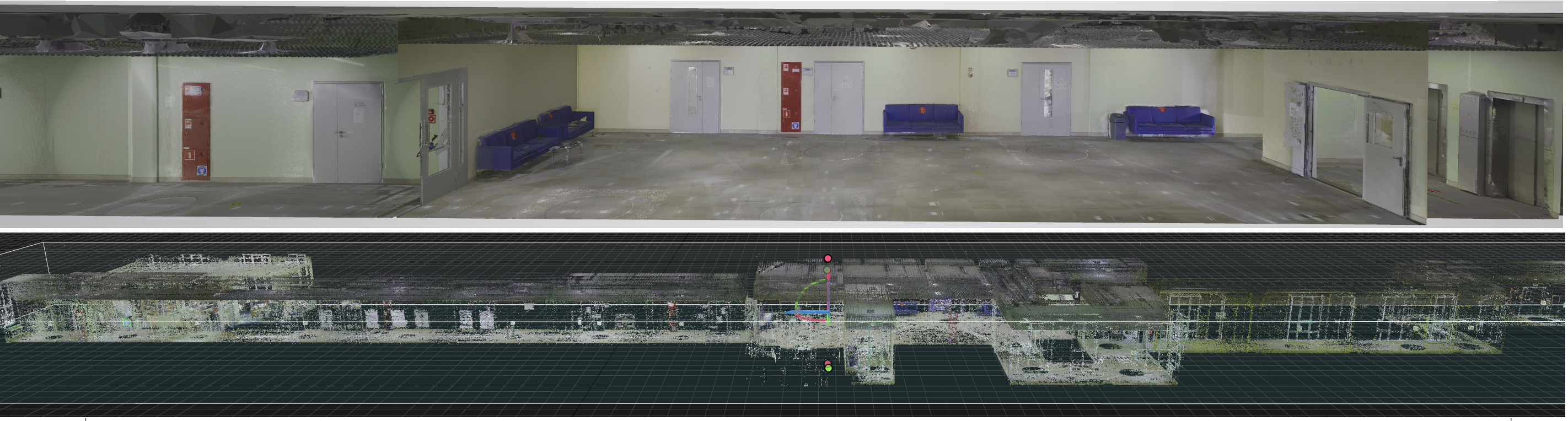}
  \caption{The bottom image is how the point cloud looks before texturing. The upper image is after texturing at RealityCapture.}
   \label{fig:mipt_env_image}
\end{figure*}

\begin{table*}[htbp]
  \centering
    \begin{tabular}{l|cccc}
        \hline
        \multirow{2}{*}{\bfseries Parameter}&
        \multicolumn{3}{c}{\bfseries PPO Parameters}\\
        
        & {\bfseries PointNav} & {\bfseries Explore} & {\bfseries GoalReacher}\\ \hline
        Visual input shape         & RGBD: (4,160,120) & RGBD: (4,320,240) & Depth: (1,128,96)\\
        CNN backbone                & Resnet9 & Resnet9 & Resnet9\\
        RNN type                    & LSTM & GRU & LSTM\\
        Number of RNN layers         & 2 & 2 & 1\\
        PPO Mini-Batches                & 2 & 2 & 2\\
        PPO Clip                    & 0.2 & 0.2 & 0.2\\
        $\gamma$                 & 0.99 & 0.99 & 0.99\\
        GAE                   & 0.95 & 0.95 & 0.95\\
        Learning rate                   & 0.00025 & 0.00025 & 0.00025\\
        Number of environments  & 22 & 20 & 28\\
        Rollout length       & 32 & 32 & 32\\
        \hline 
    \end{tabular}
  \caption{Hyper-parameters used for training skills.}
  \label{tab:hyperparameters}
\end{table*}

\newpage

All skill policies in our method are using RL. A policy ${\pi}$ of the agent is a function that maps a state to a distribution over actions. De facto, the policy and the value functions are represented as two neural networks. The first represents the current policy $\pi$, and a value network approximates the current policy's value function ${V}\approx{V}^\pi$.
A policy neural network is a two-head network, one for the action distribution (actor) and the other for the action value estimation (critic). The actor stream is one FC layer that outputs logits for each out of four actions. An action to execute is picked as a categorical distribution of that logits. The critic stream is an FC layer that outputs value for the given state. The policy hyper-parameters are listed (Table 1) in supplementary materials.